\documentclass[10pt,twocolumn,letterpaper]{article}

\usepackage[pagenumbers]{cvpr} 

\usepackage{graphicx}
\usepackage{amsmath}
\usepackage{amssymb}
\usepackage{booktabs}
\usepackage{algorithm}
\usepackage{algorithmic}
\usepackage{tabularx}
\usepackage{multirow}
\usepackage{hhline}
\usepackage[table,dvipsnames]{xcolor}
\usepackage{enumitem}

\makeatletter
\def\thickhline{%
  \noalign{\ifnum0=`}\fi\hrule \@height \thickarrayrulewidth \futurelet
  \reserved@a\@xthickhline}
\def\@xthickhline{\ifx\reserved@a\thickhline
              \vskip\doublerulesep
              \vskip-\thickarrayrulewidth
             \fi
      \ifnum0=`{\fi}}
\makeatother
\newlength{\thickarrayrulewidth}
\newcolumntype{Y}{>{\centering\arraybackslash}X}

\usepackage[pagebackref,breaklinks,colorlinks]{hyperref}

\usepackage[capitalize]{cleveref}
\crefname{section}{Sec.}{Secs.}
\Crefname{section}{Section}{Sections}
\Crefname{table}{Table}{Tables}
\crefname{table}{Tab.}{Tabs.}

\def\cvprPaperID{13} 
\def\confName{CVPR}
\def\confYear{2022}

\begin{document}

\title{SeCGAN: Parallel Conditional Generative Adversarial Networks for Face Editing via Semantic Consistency}

\author{Jiaze Sun$^1$\thanks{Corresponding author (j.sun19@imperial.ac.uk).} \quad Binod Bhattarai$^{2}$\thanks{Part of work done while at Imperial College London.} \quad Zhixiang Chen$^{3}$\footnotemark[2] \quad Tae-Kyun Kim$^{1,4}$ \\
{\small $^1$Imperial College London \quad $^2$University College London \quad $^3$University of Sheffield \quad $^4$Korea Advanced Institute of Science and Technology} \\
}


\maketitle

\begin{abstract}
Semantically guided conditional Generative Adversarial Networks (cGANs) have become a popular approach for face editing in recent years.
However, most existing methods introduce semantic masks as direct conditional inputs to the generator and often require the target masks to perform the corresponding translation in the RGB space.
We propose SeCGAN, a novel label-guided cGAN for editing face images utilising semantic information without the need to specify target semantic masks.
During training, SeCGAN has two branches of generators and discriminators operating in parallel, with one trained to translate RGB images and the other for semantic masks.
To bridge the two branches in a mutually beneficial manner, we introduce a semantic consistency loss which constrains both branches to have consistent semantic outputs.
Whilst both branches are required during training, the RGB branch is our primary network and the semantic branch is not needed for inference.
Our results on CelebA, CelebA-HQ, and RaFD demonstrate that our approach is able to generate facial images with more precise attributes/expressions, outperforming competitive baselines in terms of editing accuracy whilst maintaining quality metrics such as self-supervised Fr\'{e}chet Inception Distance and Inception Score.
\end{abstract}

\section{Introduction}


\begin{figure}
    \centering
    \includegraphics[width=0.99\linewidth]{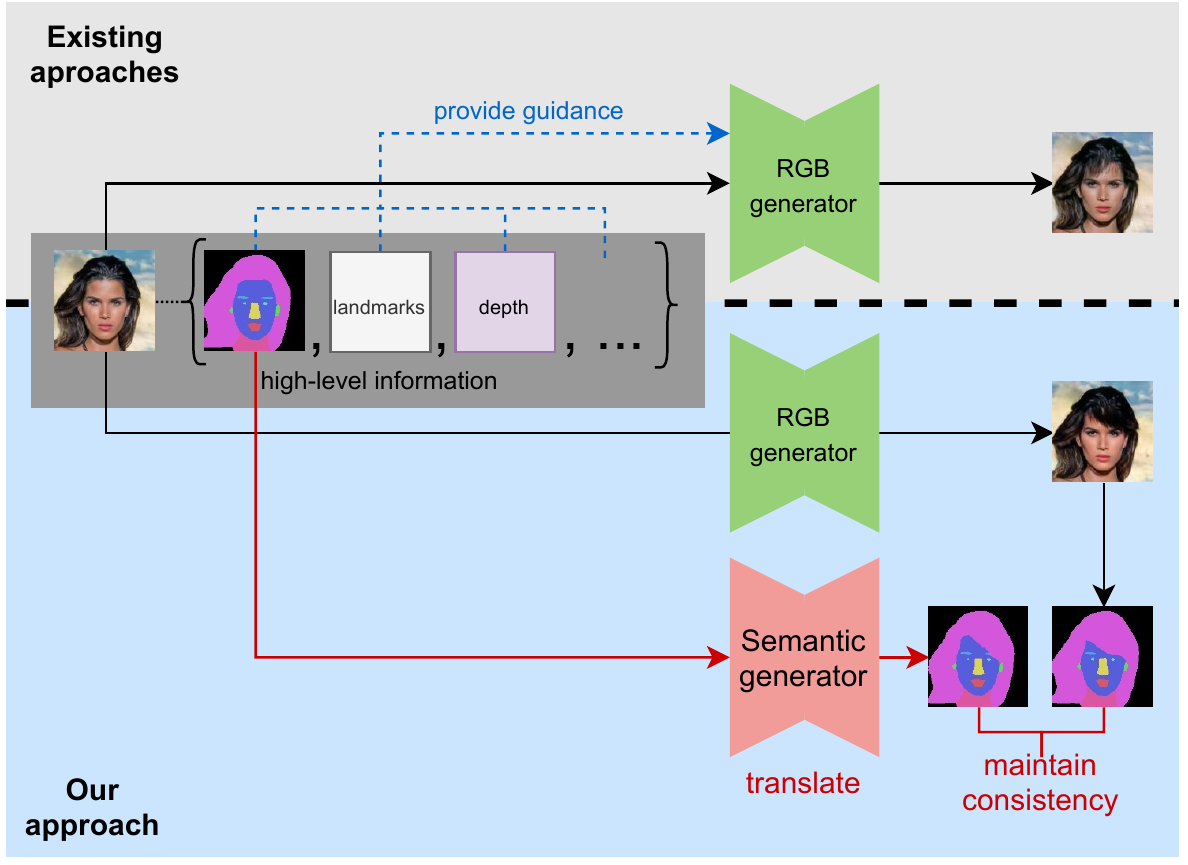}
    \caption{Existing approaches usually incorporate high-level information within the RGB domain. We propose to translate semantic masks in parallel to the RGB images without altering existing the architecture in the RGB domain (zoom in for a better view).}
    \label{fig:conceptual_difference}
\end{figure}

The advent of deep learning~\cite{LeCun1989Backpropagation,He2016Deep,Schroff2015FaceNet} led to a growing demand for annotated training data. However, it is challenging in many practical tasks to collect large-scale annotated data as it demands much expertise, time, and resources. This becomes even more difficult for face-related tasks due to privacy concerns. To tackle this, target-labelled conditional Generative Adversarial Networks (cGANs) have been widely used to generate photo-realistic face images or edit existing images to produce additional labelled ones.

Generating such examples is an important and challenging research problem~\cite{choi2018stargan,He2019AttGAN,Liu2019STGAN,bhattarai2020inducing,Sun2020MatchGAN}. In past years, attempts were made focusing on encoding target domain information in order to regularise cGANs and improve the accuracy and quality of generated images. A number of them focused on engineering representations of target labels and have managed to improve the performance. STGAN~\cite{Liu2019STGAN} proposed to condition the generator using the difference between binary vector representations of the target and source labels~\cite{Liu2019STGAN} instead of target only~\cite{choi2018stargan,mirza2014conditionalgan}. A more recent study on cGAN conditioning~\cite{bhattarai2020inducing} observes that higher-level encodings such as word embeddings and graph-induced representations are more effective in representing target information compared to traditional forms such as binary vectors. In addition to label encoding, geometry-based higher-level representations such as facial semantic masks~\cite{Gu2019Mask,Lee2020MaskGAN}, 3D morphable model (3DMM) parameters for faces~\cite{gecer2018semi}, and sparse and dense facial key points~\cite{Qian2019Make,kossaifi2018gagan} have also been employed to encode target information and provide guidance for cGANs. In comparison to methods relying only on low level RGB images~\cite{choi2018stargan,He2019AttGAN}, these geometry-based methods have been shown to provide more effective and accurate guidance for cGANs. However, a commonality between these methods is that the generation output stays within the RGB domain and the higher-level information is either absorbed through direct input or in a manner to supplement the RGB data.

\begin{figure*}
\begin{center}
\includegraphics[width=0.95\linewidth]{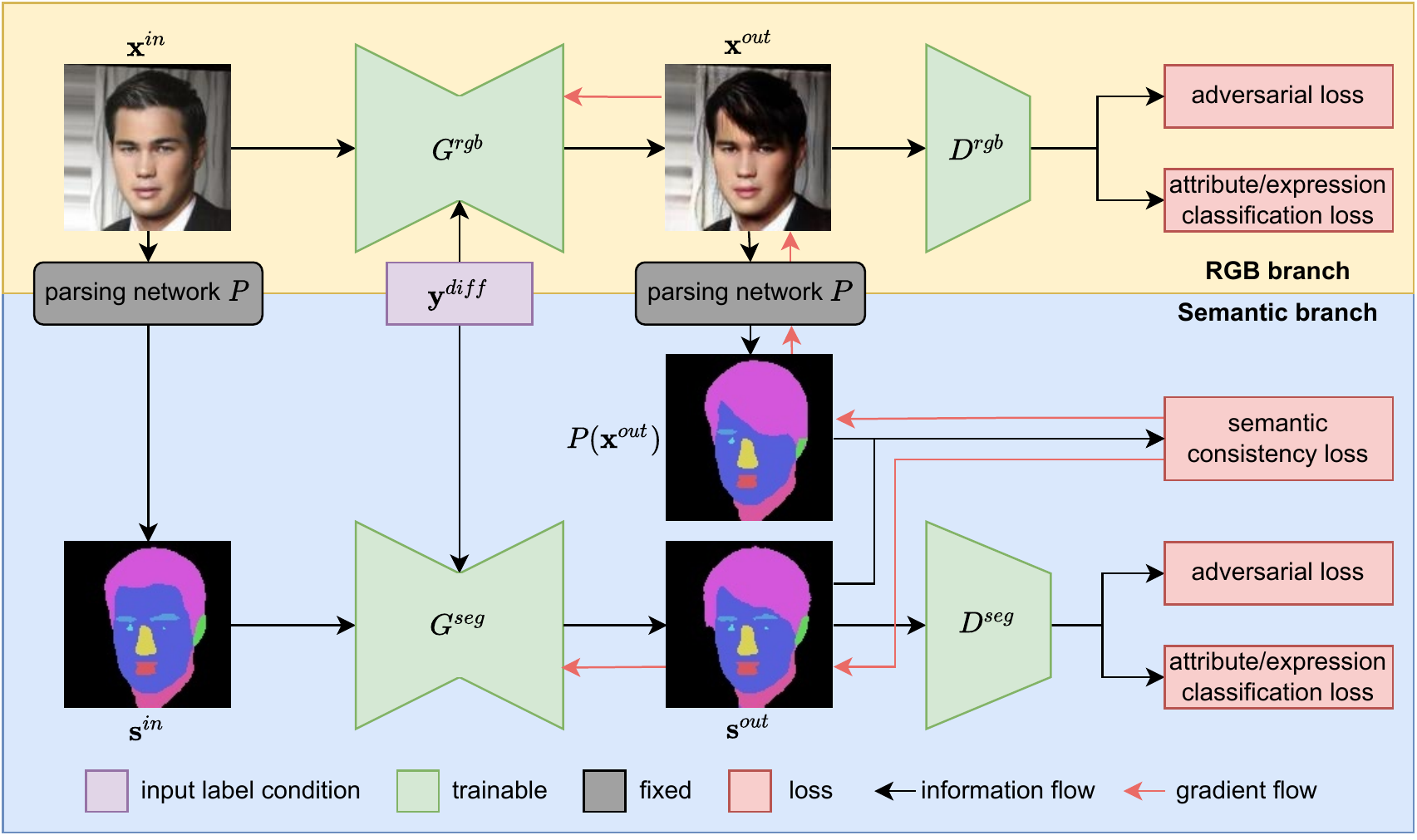}
\end{center}
   \caption{\textbf{Overall pipeline of our method.} The input image $\mathbf{x}^{in}$ is translated to $\mathbf{x}^{out}$ by $G^{rgb}$ in the RGB branch. Meanwhile, $\mathbf{x}^{in}$ is parsed by the segmentation network $P$ to obtain the mask $\mathbf{s}^{in}$ which is translated to $\mathbf{s}^{out}$ by $G^{seg}$ in the semantic branch. The translated RGB image $\mathbf{x}^{out}$ is further parsed by $P$. Inconsistency between the two semantic outputs is measured by the semantic consistency loss and minimised by back-propagating gradient into $G^{rgb}$ and $G^{seg}$. For clarity, the reconstruction loss is not shown in this figure.}
\label{fig:archi}
\end{figure*}

We hypothesise that, by distributing high and low level information to separate network branches and training them in a cooperative manner, the overall model would be able to utilise different modalities more effectively than by simply absorbing the high-level information via a direct input. Unlike previous methods which take higher-level information simply to augment or fuse with the RGB data \cite{Gu2019Mask,Tang2020XingGAN}, we propose \emph{translating} higher-level facial descriptions and the RGB images in parallel and imposing certain higher-level consistency constraint in the target domain (shown in Figure~\ref{fig:conceptual_difference}). To this end, we propose to train a semantic mask translation network in parallel to the RGB network as shown in Figure~\ref{fig:archi}. We choose semantic mask for higher-level information as it is denser and more relevant to our task compared to other choices such as landmarks. However, the proposed framework can easily be extended to other such descriptions of the face and implemented on top of existing frameworks without altering their architecture or overriding their existing capacity. Unlike previous work employing semantic masks \cite{Gu2019Mask,Lee2020MaskGAN} which require target mask annotations, we only have access to the source mask and thus our work is orthogonal to these methods.
In addition, our framework also benefits from adopting a hierarchical structure which has been shown to retain target attributes better by preventing the modification of unwanted attributes~\cite{yi2019apdrawinggan,Wu2020Cascade}. Compared to existing hierarchical cGANs for face editing~\cite{Wu2020Cascade} whose hierarchy is based on spatial locations and bounding boxes, our method is based on semantic segmentation components which are more refined and allow for more precise manipulations.

We would like to emphasise that our main goal in this work is to utilise semantic information to generate images with \emph{greater accuracy} whilst \emph{maintaining image quality}. In summary, our contributions are as follows:
\begin{itemize} 
    \item We present a cGAN for face editing, employing a parallel hierarchy to perform translation at both raw-pixel and semantic level. To our knowledge, this is the first work exploring image-to-image translation purely between semantic masks.
    \item Our extensive quantitative evaluations show that our method achieves \emph{superior} editing accuracy whilst able to \emph{maintain} quality metrics including self-supervised Fr\'{e}chet Inception Distance (ssFID) and Inception Score (IS) across multiple datasets in the same domain.
    \item Our qualitative evaluations show that our method is able to synthesise more distinct and accurate attributes whilst avoiding unnecessary and irrelevant edits.
\end{itemize}
\section{Related work}
\textbf{Image-to-image translation with cGANs.} Label-conditioned GANs \cite{mirza2014conditionalgan} are widely used in image-to-image translation tasks. They take a label vector as an additional input to help guide the translation process thus providing better control over the output images. Pix2pix \cite{phillip2016image} and CycleGAN \cite{zhu2017cyclegan} were amongst the first such frameworks, but they must be trained separately for each pair of domains. StarGAN \cite{choi2018stargan} and AttGAN \cite{He2019AttGAN} both overcame this challenge and are able to perform multi-domain translations using a single generator conditioned on the target label. STGAN \cite{Liu2019STGAN} improved AttGAN by conditioning the generator using the difference between the target and source labels and incorporating selective transfer units. \cite{bhattarai2020inducing} further improved STGAN by transforming the attribute difference vector with a graph neural network before using it to condition the generator. These methods all rely purely on the RGB domain without any higher-level geometric information.

\textbf{Geometry-guided generation.} Geometric information has also been incorporated into various GAN architectures to guide the generation process. \cite{Gu2019Mask,Lee2020MaskGAN} proposed using source and target semantic segmentation masks as direct inputs to the generator for face style transfer and component editing. \cite{Qian2019Make} used facial landmarks to manipulate facial expressions and head poses. \cite{men2020controllable} uses both pose and semantic information to synthesise whole body images. \cite{Jo2019SCFEGAN} edits facial images using input sketches as reference. \cite{gecer2018semi} uses 3DMM to generate identity and pose before adding realism using a GAN. Most work in this area requires a ground-truth or user-specified target semantic mask as a reference for generation. In our work, the only guidance is the information provided by the source label and semantic masks and we use a GAN to generate the target mask.

\textbf{Hierarchical GANs and mutual learning.} Most frameworks mentioned above feature a single GAN in their architecture, but recent work has started to feature multiple GANs operating in different pathways, with each operating on a different scale or modality. SinGAN \cite{Shaham2019SinGAN} can generate high-resolution images using a pyramid of GANs, with generators at higher-resolutions taking information from those at lower resolutions. \cite{yi2019apdrawinggan} translates facial images to artistic drawings by employing a global and multiple local GANs that focus on individual facial components before fusing them together. \cite{Wu2020Cascade} learns to manipulate facial expressions via a similar global-local hierarchy. \cite{Chen2019Learning} adds realism to synthetic 3D images via two adversarial games - one in the RGB domain and the other in the semantic and depth domain. \cite{Tang2020XingGAN} performs pose editing by splitting the generator into an RGB branch and pose branch which share latent information with each other. Inspired by \cite{Chen2019Learning} and \cite{Tang2020XingGAN}, we propose to incorporate two branches of GANs performing translations in separate domains, RGB and semantic, whilst sharing information in a mutual learning manner \cite{Zhang2018Deep}.

\section{Methodology}
In this section, we present in detail our approach for face editing via semantic consistency. Given an input face image $\mathbf{x}^{in}$ with source attribute/expression label $\mathbf{y}^{src}$ and a target one $\mathbf{y}^{trg}$, we aim to synthesise an image $\mathbf{x}^{out}$ with the desired characteristics specified by $\mathbf{y}^{trg}$. 

Our approach is to incorporate semantic masks as a parallel modality to RGB images in the training process. Unlike previous work \cite{Gu2019Mask,Lee2020MaskGAN} which require target segmentation masks to be specified as an input to their networks, we have no access to any ground truth target mask information. Instead, we propose to generate the target mask by translating the source mask using a second generator $G^{seg}$, trained in parallel with the standard generator $G^{rgb}$ used for translating RGB images. The outputs of both generators are then compared against each other at a semantic level and this information is back-propagated to both generators to minimise any semantic inconsistency between the two outputs. Whilst $G^{rgb}$ is our primary generator and will be kept during inference, $G^{seg}$ only provides auxiliary guidance during training and is not necessary for inference.

\textbf{Parsing network.} We used an existing semantic segmentation network \cite{Yu2018BiSeNet} pre-trained on CelebAMask-HQ \cite{Lee2020MaskGAN} to parse both source and translated RGB images into semantic regions. The network is pre-trained separately for different resolutions and classifies each pixel in the RGB domain into one of the following 12 segments: skin, eyebrows, eyes, eyeglasses, ears, earrings, nose, mouth, lips, neck, hair, and others. The last segment ``others'' is in fact the union of background, necklace, hat, and clothes which are irrelevant to face editing in our case.

\subsection{Parallel GANs}
The overall pipeline of our framework is shown in Figure \ref{fig:archi}. During training, two branches of generators ($G^{rgb}$, $G^{seg}$) and discriminators ($D^{rgb}$, $D^{seg}$) operate in parallel, with $G^{rgb}$ and $D^{rgb}$ forming the RGB branch and $G^{seg}$ and $D^{seg}$ forming the semantic branch. The framework of either branch can be substituted by that of a single GAN such as StarGAN \cite{choi2018stargan}, AttGAN \cite{He2019AttGAN}, or STGAN \cite{Liu2019STGAN}.

\textbf{RGB branch.} Given an input RGB image $\mathbf{x}^{in}$, source and target labels $\mathbf{y}^{src}$ and $\mathbf{y}^{trg}$, $G^{rgb}$ synthesises an output RGB image $\mathbf{x}^{out} = G^{rgb}\left(\mathbf{x}^{in}, \mathbf{y}^{diff}\right)$, where $\mathbf{y}^{diff} = \mathbf{y}^{trg} - \mathbf{y}^{src}$ following STGAN \cite{Liu2019STGAN}. The discriminator $D^{rgb}$ takes one image as input and has two output heads, $D_{adv}^{rgb}$ and $D_{cls}^{rgb}$, corresponding to predictions of realism and attributes/expressions respectively. We train the network to produce realistic images using the WGAN-GP loss \cite{Gulrajani2017Improved}
\begin{align}
    \begin{split}
        \mathcal{L}_{adv}^{D^{rgb}} = & \mathop{\mathbb{E}}_{\substack{\mathbf{x}^{in},\mathbf{y}^{diff}}}[D^{rgb}(G^{rgb}(\mathbf{x}^{in},\mathbf{y}^{diff}))] \\
                                    & - \mathop{\mathbb{E}}_{\mathbf{x}^{in}}[D^{rgb}(\mathbf{x}^{in})] + \lambda_{gp}\mathop{\mathbb{E}}_{\tilde{\mathbf{x}}}[\|\nabla_{\tilde{\mathbf{x}}}D^{rgb}(\tilde{\mathbf{x}})\|_2-1]^2,
    \end{split}\\
    \mathcal{L}_{adv}^{G^{rgb}} = & \mathop{\mathbb{E}}_{\mathbf{x}^{in},\mathbf{y}^{diff}}[-D^{rgb}(G^{rgb}(\mathbf{x}^{in},\mathbf{y}^{diff}))],
\end{align}
where $\tilde{\mathbf{x}}$ is uniformly sampled along straight lines between $\mathbf{x}^{in}$ and $\mathbf{x}^{out}$. To generate the desired target attributes/expressions, we minimise the classification loss
\begin{align}
    \begin{split}
        \mathcal{L}_{cls}^{D^{rgb}} &= \mathop{\mathbb{E}}_{\mathbf{x}^{in},\mathbf{y}^{src}}-[\mathbf{y}^{src}\cdot\log(D_{cls}^{rgb}(\mathbf{x}^{in}))\\
                                    &+ (\mathbf{1} - \mathbf{y}^{src})\cdot(\mathbf{1} - \log(D_{cls}^{rgb}(\mathbf{x}^{in})))],
    \end{split}\\
    \begin{split}
        \mathcal{L}_{cls}^{G^{rgb}} &= \mathop{\mathbb{E}}_{\mathbf{x}^{in},\mathbf{y}^{trg}}-[\mathbf{y}^{trg}\cdot\log(D_{cls}^{rgb}(G^{rgb}(\mathbf{x}^{in},\mathbf{y}^{diff}))) \\
                                &+ (\mathbf{1} - \mathbf{y}^{trg})\cdot(\mathbf{1} - \log(D_{cls}^{rgb}(G^{rgb}(\mathbf{x}^{in},\mathbf{y}^{diff}))))],
    \end{split}
\end{align}
where ``$\cdot$'' denotes the dot product, $\log$ is element-wise, and $\mathbf{1}$ is the all-one vector. Finally, to ensure that the translated images are consistent with the input images, we impose the reconstruction loss
\begin{equation}
    \mathcal{L}_{rec}^{rgb} = \mathop{\mathbb{E}}_{\mathbf{x}^{in}} [\|\mathbf{x}^{in}-G^{rgb}(\mathbf{x}^{in},\mathbf{0})\|_1],
    \label{eqn:rec_rgb}
\end{equation}
where $\mathbf{0}$ is the zero vector and $\|\cdot\|_1$ is the L1 norm.

\textbf{Semantic branch.} The semantic branch operates in a manner similar to the RGB branch and the main difference is that the former only translates semantic segmentation masks. Given an input RGB image $\mathbf{x}^{in}$, we generate its soft semantic mask $\hat{\mathbf{s}}^{in}=P(\mathbf{x}^{in})$ using the aforementioned fixed parsing network $P$. Here, the soft mask $\hat{\mathbf{s}}^{in}$ is further represented in a one-hot fashion, $\mathbf{s}^{in}$, which is of shape [number of segments $\times$ height $\times$ width] and each pixel has only one channel equal to 1 and all others 0. Similar to the RGB branch, given $\mathbf{y}^{src}$ and $\mathbf{y}^{trg}$, the semantic generator $G^{seg}$ synthesises an output semantic mask $\hat{\mathbf{s}}^{out} = G^{seg}\left(\mathbf{s}^{in}, \mathbf{y}^{diff}\right)$ which has the same shape as the input mask. Rather than using a $\tanh$ layer as the final activation before the output (which is the case for the RGB branch), we replace that with a Softmax layer so that each output pixel is a probability distribution over the 12 segments and that gradient is able to flow through during back-propagation. On the other hand, the discriminator $D^{seg}$ functions largely the same way as its RGB counterpart, taking a semantic mask as input and outputting predictions for the realism score $D_{adv}^{seg}$ and label vector $D_{cls}^{seg}$. As for the loss functions, we also train the semantic branch using the WGAN-GP adversarial loss
\begin{align}
    \begin{split}
        \mathcal{L}_{adv}^{D^{seg}} = & \mathop{\mathbb{E}}_{\mathbf{s}^{in},\mathbf{y}^{diff}}[D^{seg}(G^{seg}(\mathbf{s}^{in},\mathbf{y}^{diff}))] \\
                                  & - \mathop{\mathbb{E}}_{\mathbf{s}^{in}}[D^{seg}(\mathbf{s}^{in})] + \lambda_{gp}\mathop{\mathbb{E}}_{\tilde{\mathbf{s}}}[\|\nabla_{\tilde{\mathbf{s}}}D^{seg}(\tilde{\mathbf{s}})\|_2-1]^2,
    \end{split}\\
    \mathcal{L}_{adv}^{G^{seg}} = & \mathop{\mathbb{E}}_{\mathbf{s}^{in},\mathbf{y}^{diff}}[-D^{seg}(G^{seg}(\mathbf{s}^{in},\mathbf{y}^{diff}))],
\end{align}
the attribute/expression classification loss
\begin{align}
    \begin{split}
        \mathcal{L}_{cls}^{D^{seg}} &= \mathop{\mathbb{E}}_{\mathbf{s}^{in},\mathbf{y}^{src}}-[\mathbf{y}^{src}\cdot\log(D_{cls}^{seg}(\mathbf{s}^{in}))\\
                                    &+ (\mathbf{1} - \mathbf{y}^{src})\cdot(\mathbf{1} - \log(D_{cls}^{seg}(\mathbf{s}^{in})))],
    \end{split}\\
    \begin{split}
        \mathcal{L}_{cls}^{G^{seg}} &= \mathop{\mathbb{E}}_{\mathbf{s}^{in},\mathbf{y}^{trg}}-[\mathbf{y}^{trg}\cdot\log(D_{cls}^{seg}(G^{seg}(\mathbf{s}^{in},\mathbf{y}^{diff}))) \\
                                &+ (\mathbf{1} - \mathbf{y}^{trg})\cdot(\mathbf{1} - \log(D_{cls}^{seg}(G^{seg}(\mathbf{s}^{in},\mathbf{y}^{diff}))))],
    \end{split}
\end{align}
and the reconstruction loss
\begin{equation}
    \mathcal{L}_{rec}^{seg} = \mathop{\mathbb{E}}_{\mathbf{s}^{in}} \left[-\frac{1}{HW}\sum_{i,j}\mathbf{s}_{i,j}^{in}\cdot \log\left(G^{seg}(\mathbf{s}^{in},\mathbf{0})_{i,j}\right)\right],
    \label{eqn:rec_seg}
\end{equation}
where $i,j$ denotes the $ij$-th pixel, and $H$ and $W$ are the image height and width respectively. Unlike the RGB branch which uses the L1 loss for reconstruction, we use the cross entropy loss here as it is more suitable for optimising probability distributions.

\subsection{Semantic consistency}
Up to this point, the two branches of GANs are still independent from each other. To link both networks so that they can mutually benefit from each other during training, we introduce the semantic consistency loss as an additional term to the overall objective. Unlike knowledge distillation frameworks which usually involve a one-way information transfer from a large (teacher) network to a small (student) one, our method is more akin to a mutual learning objective \cite{Zhang2018Deep} as complementary information can be transferred both ways between $G^{rgb}$ and $G^{seg}$ during training. While $G^{rgb}$ focuses on translations at the raw pixel level, it can easily neglect higher order information such as the geometry of facial components, but this can be compensated by the $G^{seg}$ network. On the other hand, $G^{seg}$ operating purely at the semantic level might lead to drastic changes in the translated images which might not be desirable in the RGB domain, and thus can be held in check by the RGB branch.

More specifically, we impose the semantic consistency loss on the soft semantic outputs of the two generators, namely $P(\mathbf{x}^{out})$ and $\hat{\mathbf{s}}^{out}$. Since these are essentially classification probabilities, we choose to measure the discrepancy between $P(\mathbf{x}^{out})$ and $\hat{\mathbf{s}}^{out}$ using the cross entropy loss. To update $G^{rgb}$, we minimise the following loss
\begin{align}
    \mathcal{L}_{sc}^{G^{rgb}} &= \mathop{\mathbb{E}}_{\mathbf{s}^{out},\mathbf{x}^{out}} \left[-\frac{1}{HW}\sum_{i,j}\mathbf{s}_{i,j}^{out}\cdot \log\left(P(\mathbf{x}^{out})_{i,j})\right)\right],
\end{align}
where $\mathbf{s}^{out}$ is the one-hot representation of $\hat{\mathbf{s}}^{out}$. Similarly, to update $G^{seg}$, we minimise
\begin{align}
    \mathcal{L}_{sc}^{G^{seg}} &= \mathop{\mathbb{E}}_{\hat{\mathbf{s}}^{out},\mathbf{x}^{out}} \left[-\frac{1}{HW}\sum_{i,j}\bar{P}(\mathbf{x}^{out})_{i,j}\cdot \log\left(\hat{\mathbf{s}}_{i,j}^{out})\right)\right],
\end{align}
where $\bar{P}(\mathbf{x}^{out})$ is the one-hot representation of $P(\mathbf{x}^{out})$. This way, both $G^{rgb}$ and $G^{seg}$ learn to generate outputs with consistent semantic structures.

\subsection{Optimisation}
\textbf{Overall loss functions.} Summarising all the loss terms we have introduced so far, we have
\begin{align}
    \mathcal{L}^{D^{rgb}} &= \mathcal{L}_{adv}^{D^{rgb}} + \lambda_{cls}\mathcal{L}_{cls}^{D^{rgb}}, \label{eqn:d_rgb} \\
    \mathcal{L}^{G^{rgb}} &= \mathcal{L}_{adv}^{G^{rgb}} + \lambda_{cls}\mathcal{L}_{cls}^{G^{rgb}} + \lambda_{rec}\mathcal{L}_{rec}^{rgb} + \lambda_{sc}\mathcal{L}_{sc}^{G^{rgb}}, \label{eqn:g_rgb} \\
    \mathcal{L}^{D^{seg}} &= \mathcal{L}_{adv}^{D^{seg}} + \lambda_{cls}\mathcal{L}_{cls}^{D^{seg}}, \label{eqn:d_seg} \\
    \mathcal{L}^{G^{seg}} &= \mathcal{L}_{adv}^{G^{seg}} + \lambda_{cls}\mathcal{L}_{cls}^{G^{seg}} + \lambda_{rec}\mathcal{L}_{rec}^{seg} + \lambda_{sc}\mathcal{L}_{sc}^{G^{seg}}. \label{eqn:g_seg}
\end{align}

\textbf{Training.} In a fashion similar to \cite{Zhang2018Deep}, both the RGB and semantic branches receive the same mini-batch of examples during each iteration, and the semantic consistency loss is used to update the parameters of one generator based on the semantic output of the other. The details of the training procedure is summarised in Algorithm \ref{alg:training} in Appendix \ref{sec:algorithm}.
\section{Experiments}

\subsection{Datasets}
\textbf{CelebA.} The CelebFaces Attributes Dataset (CelebA) \cite{liu2015faceattributes} consists of 202,599 facial images of size $178\times 218$, which were centre-cropped to $178\times 178$ and then resized to a desired resolution for training ($128\times 128$ or $256\times 256$) in our experiments. Each example has 40 attribute annotations, from which we selected 13 attributes for our experiments, namely bald, bangs, black hair, blonde hair, brown hair, bushy eyebrows, eyeglasses, gender, mouth slightly open, moustache, beard, pale skin, and age. We allocated 182,000 images for training, 637 for validation, and 19,962 for testing, following \cite{He2019AttGAN}. 

\textbf{CelebA-HQ.} This dataset \cite{karras2018progressive,Lee2020MaskGAN} is a subset of CelebA with 30,000 examples in total, but each image is re-created in $1024\times1024$ by \cite{karras2018progressive}. Each image also has 40 attribute annotations from which we selected the same 13 as before. We split the dataset into 28,000 for training, 500 for validation, and 1,500 for testing, again following \cite{He2019AttGAN}.

\textbf{RaFD.} The Radboud Faces Database (RaFD) \cite{Langner2010rafd} comprises facial images of 67 human models showing 8 expressions including angry, contemptuous, disgusted, fearful, happy, neutral, sad, and surprised. From each original $681\times1024$ image, we obtain a $600\times600$ crop around the centre of the head and resize it to a desired resolution for training. RaFD is a significantly smaller dataset than CelebA. We randomly select 60 models (4320 images) as training examples and the remaining 7 (504 images) as test.

\subsection{Baselines}
We adopted two baseline models for our experiments, StarGAN \cite{choi2018stargan} and AttGAN \cite{He2019AttGAN}, both of which perform label conditioned image-to-image translations.

\textbf{StarGAN.} StarGAN is a cGAN which unifies multi-domain image-to-image translation using a single generator. Whilst the generator in StarGAN does not strictly have an encode-decoder architecture, it has six residual blocks serving as bottleneck layers sandwiched between down-sampling layers on the input side and upsampling layers on the output side. The discriminator is based on PatchGAN \cite{phillip2016image}, a fully convolutional network which learns and assesses features at the scale of local image patches. In addition, StarGAN uses a cycle-consistency loss \cite{zhu2017cyclegan} given by $\mathop{\mathbb{E}}_{\mathbf{x},\mathbf{y}^{src},\mathbf{y}^{trg}} [\|\mathbf{x}-G(G(\mathbf{x},\mathbf{y}^{trg}),\mathbf{y}^{src})\|_1]$. Since we are using the label difference vector $\mathbf{y}^{diff}$ as the conditional input in this work, there is no need to perform two translations to reconstruct a given input image. Therefore, we replace it with the reconstruction loss introduced before in equations (\ref{eqn:rec_rgb}) and (\ref{eqn:rec_seg}) for our experiments.

\textbf{AttGAN.} AttGAN is a cGAN with a similar functionality and pipeline as StarGAN. Unlike StarGAN, however, AttGAN features an encode-decoder architecture which includes skip connections between them as well as injection layers where label vectors are concatenated with latent representations at various scales. Unlike StarGAN, the discriminator of AttGAN employs two fully-connected layers following a cascade of convolutional ones. Similarly to our StarGAN baseline, we also use the label difference vector $\mathbf{y}^{diff}$ as the conditional input to the generator. Compared to the 8M trainable parameters in StarGAN's generator, AttGAN's generator has significantly more at 43M and serves as a higher-capacity use case in our tests.

\subsection{Implementation details}
We implemented our method in PyTorch based on the official implementation of StarGAN and the PyTorch version of AttGAN. We follow the same training procedures and values for $\lambda_{cls}$, $\lambda_{rec}$, and $\lambda_{gp}$ as the original implementations. For StarGAN, we trained the network for 200K iterations with the generator being updated once every 5 discriminator updates, whereas for AttGAN the training lasted roughly 1.1M iterations with the same generator update frequency. We used the Adam optimiser \cite{Kingma2015Adam} for both architectures with $\beta_1=0.5$ and $\beta_2=0.999$. We set initial learning rates of $1\times10^{-4}$ and $2\times10^{-4}$ for StarGAN and AttGAN respectively, which were kept constant during the first half of the training (first 100K iterations for StarGAN and first 100 epochs for AttGAN) but decayed linearly to 0 for StarGAN and exponentially to $2\times10^{-6}$ for AttGAN. The batch size was set to 16 for StarGAN and 32 for AttGAN. On an NVIDIA RTX 2080 Ti GPU, training our framework takes about 1 day on StarGAN and 6 days on AttGAN.

\subsection{Evaluation metrics}
\textbf{Editing accuracy.} As our main task is face editing, our goal is for the model to \emph{accurately} generate the desired attributes or expressions. To this end, we evaluate the translated images using external classifiers pre-trained on the training set of CelebA and RaFD, following existing work \cite{He2019AttGAN,Liu2019STGAN,bhattarai2020inducing,Sun2020MatchGAN}. Specifically, we translate all test images by editing each of the 13 attributes or 8 expressions and measure the percentage of images that can be correctly classified by the external classifier. We report the accuracy on each attribute/expression as well as the overall mean.

\textbf{ssFID and IS.} To measure the \emph{quality} of generated images, we use the self-supervised Fr\'{e}chet Inception Distance (ssFID)~\cite{morozov2021on} and Inception Score (IS)~\cite{salimans2016improved}. The original FID~\cite{heusel2017gans} measures the \emph{similarity} between the distributions of real and fake examples by comparing their Inception-v3 embeddings. However, Inception-v3 is pre-trained by classifying ImageNet \cite{Russakovsky2015ImageNet} data and thus might not transfer well to non-ImageNet tasks or domains~\cite{morozov2021on}. The ssFID instead uses the self-supervised image representation model SwAV~\cite{Caron2020Unsupervised} to provide a more reasonable and universal assessment of image quality. The IS measures the \emph{diversity} and \emph{meaningfulness} of the generated images,
but it does not use the distribution of real images as reference. High image quality is reflected by a low ssFID and high IS. To compute the ssFID and IS, we again translate all test images by editing each of the 13 attributes or 8 expressions. For ssFID, we first compute the individual ssFIDs between the real images and each group of translated ones corresponding to a particular attribute/expression edit before taking their average. As for IS, we divide the entire pool of translated images into 10 subsets and compute the average result.

\subsection{Ablation studies}
\textbf{Direct concatenation.} To validate the advantage of incorporating a parallel semantic generator over a direct fusion of RGB and semantic masks, we evaluated the performance of an additional ``Baseline + Concat" model which takes the concatenated input of image $\mathbf{x}^{in}$ and its semantic mask $\mathbf{s}^{in}$ along the channel dimension. This way, the semantic information is made available to the generator without modifying its architecture or losses. We implemented this baseline on both StarGAN and AttGAN and the results are shown in Table \ref{tab:quantitative}. In terms of accuracy, ``Baseline + Concat" is unable to outperform the baseline and even underperforms on AttGAN. This suggests that it might be harder for a single network to handle multiple modalities effectively. Whilst ssFID appears slightly lower compared to the baseline, this difference is small and possibly due to the input mask $\mathbf{s}^{in}$ constraining the translation output to resemble the input more closely -- which could be counterproductive in face editing as reflected by accuracy.


\begin{figure}[t]
    \centering
    \includegraphics[width=\linewidth]{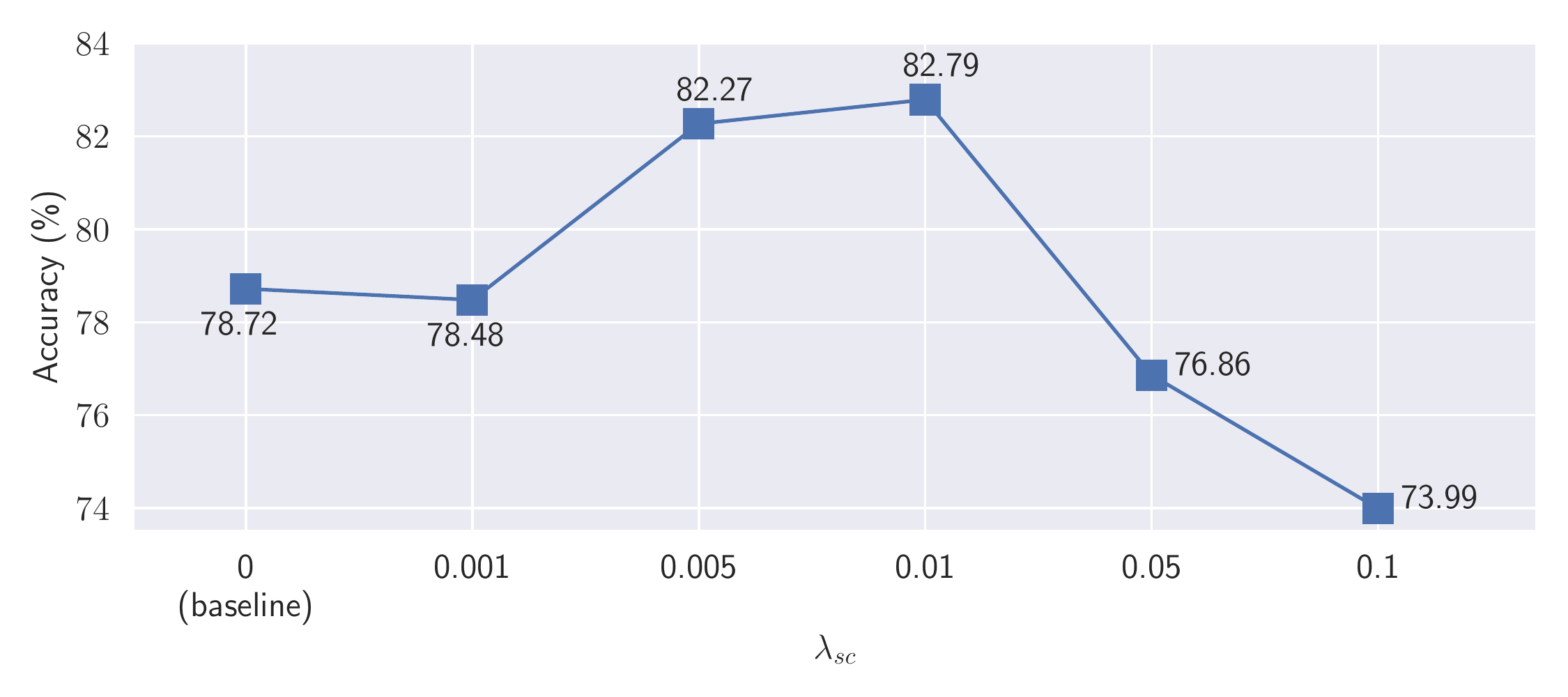}
    \caption{\bf Accuracy by loss weight parameter $\lambda_{sc}$ on CelebA.}
    \label{fig:ablation}
\end{figure}

\textbf{Consistency strength.} To maximise the potential of our method and especially accuracy, which is our main focus, we tested various values for the hyperparameter $\lambda_{sc}$ and computed the accuracy for each setup. As shown in Figure \ref{fig:ablation}, with a small $\lambda_{sc}$, the accuracy differs very little from the baseline, which indicates that the two branches are unable to exert sufficient influence over each other. The accuracy attains a maximum value of $82.79\%$ as $\lambda_{sc}$ is increased to 0.01 before starting to decrease as $\lambda_{sc}$ is increased even further. It would appear that an overly strong constraint on semantic consistency is counterproductive as it might be disruptive to the balance between various existing loss terms both across and within each of the two branches.

\subsection{Quantitative results}
\begin{table*}[!t]
    \centering
    \begin{tabularx}{\textwidth}{c|c|c|Y|>{\columncolor{Green!15}}c|c|c}
        \thickhline
        \textit{Dataset} & \textit{Resolution} & \textit{Backbone} & \textit{Method} & \textit{Accuracy (\%)} $\uparrow$ & \textit{ssFID} $\downarrow$ & \textit{IS} $\uparrow$ \\ \hline
        \multirow{6}{*}{CelebA} & \multirow{6}{*}{$128\times 128$} & \multirow{3}{*}{StarGAN} & Baseline & 78.72 & 1.58 & \textbf{3.08} \\ \hhline{~|~|~|----} 
        & & & Baseline + Concat & 80.50 & \textbf{1.47} & 3.04 \\ \hhline{~|~|~|----} 
        & & & SeCGAN (ours) & \textbf{82.79} & 1.61 & 3.07 \\ \hhline{~|~|=====}
        & & \multirow{3}{*}{AttGAN} & Baseline & 82.85 & 1.22 & 2.94 \\ \hhline{~|~|~|----} 
        & & & Baseline + Concat & 79.74 & \textbf{1.20} & \textbf{3.06} \\ \hhline{~|~|~|----} 
        & & & SeCGAN (ours) & \textbf{84.81} & 1.27 & 3.05 \\ \hhline{=======}
        \multirow{4}{*}{CelebA-HQ} & \multirow{2}{*}{$128\times128$} & \multirow{4}{*}{StarGAN} & Baseline & 72.75 & \textbf{3.22} & 2.74 \\ \hhline{~|~|~|----} 
        & & & SeCGAN (ours) & \textbf{75.78} & 3.51 & \textbf{2.80} \\ \hhline{~|-|~|====}
        & \multirow{2}{*}{$256\times 256$} & & Baseline & 58.16 & 4.97 & 3.11 \\ \hhline{~|~|~|----} 
        & & & SeCGAN (ours) & \textbf{60.17} & \textbf{4.91} & \textbf{3.12} \\ \hhline{=======}
        \multirow{2}{*}{RaFD} & \multirow{2}{*}{$128\times 128$} & \multirow{2}{*}{StarGAN} & Baseline & 86.09 & \textbf{0.92} & \textbf{1.42} \\ \hhline{~|~|~|----}
        & & & SeCGAN (ours) & \textbf{91.62} & 0.94 & 1.41 \\ \thickhline
    \end{tabularx}
    \caption{\textbf{Baselines vs our method in terms of accuracy, ssFID, and IS}. The accuracy shown here is averaged across attributes/expressions. Our method outperforms baselines consistently in terms of accuracy and is closely on par with them in terms of ssFID and IS.}
    \label{tab:quantitative}
\end{table*}

\begin{figure}
    \centering
    \includegraphics[width=\linewidth]{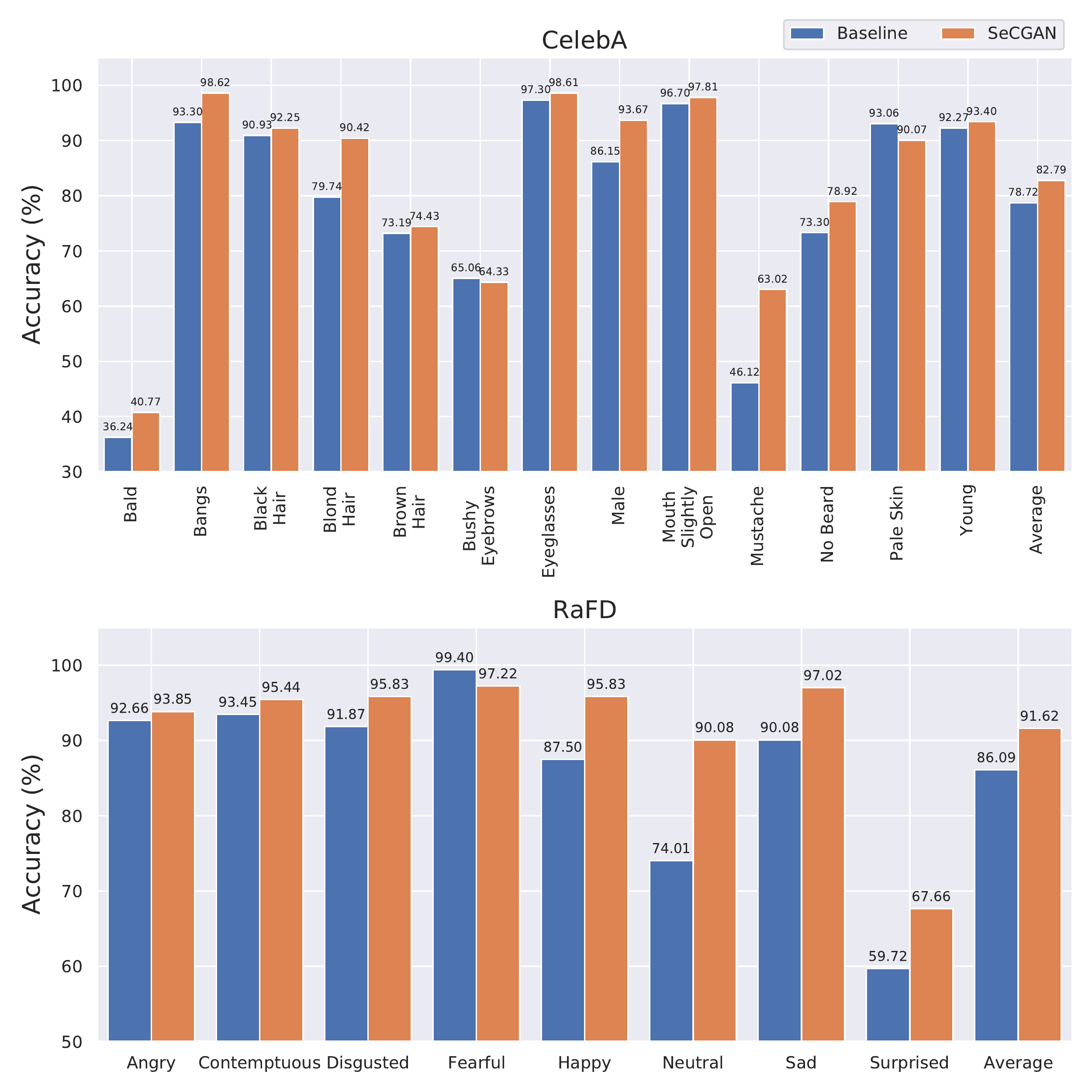}
    \caption{\textbf{Accuracy of individual attributes/expressions} (zoom in for a better view at the annotations). StarGAN is the backbone.}
    \label{fig:tarr}
\end{figure}

Table \ref{tab:quantitative} summarises the performance of the baselines and our method across three datasets, two architectures, and two resolutions. In terms of accuracy, our method consistently outperforms the baseline across all settings.
In addition, we also provide a breakdown of accuracy values by each individual attribute/expression in Figure \ref{fig:tarr}. We can also observe that our method outperforms the baseline in most of the attributes/expressions tested. This indicates that our method can indeed guide the RGB-generator towards generating more accurate attributes/expressions thanks to the higher-level information provided by the semantic branch. Table \ref{tab:quantitative} also shows that our method outperforms the baseline on RaFD, despite the parsing network not being pre-trained on the same dataset. This suggests that our method could be directly trained on new datasets without having to re-train the parsing network or require ground-truth semantic masks as long as the images used to train the parsing network and the GAN belong to the same domain.

As for image quality metrics, our method is closely on par with the baseline despite not gaining a significant advantage in every single case. In terms of IS, our method outperforms the baseline in most of the settings, indicating that the diversity and meaningfulness of our generated images are at least as good as the baseline. In terms of ssFID, our method is roughly the same as the baseline which shows that the image quality is well maintained and not adversely impacted by the gain in other metrics.

\subsection{Qualitative results}
\begin{figure*}
    \centering
    \includegraphics[width=\textwidth]{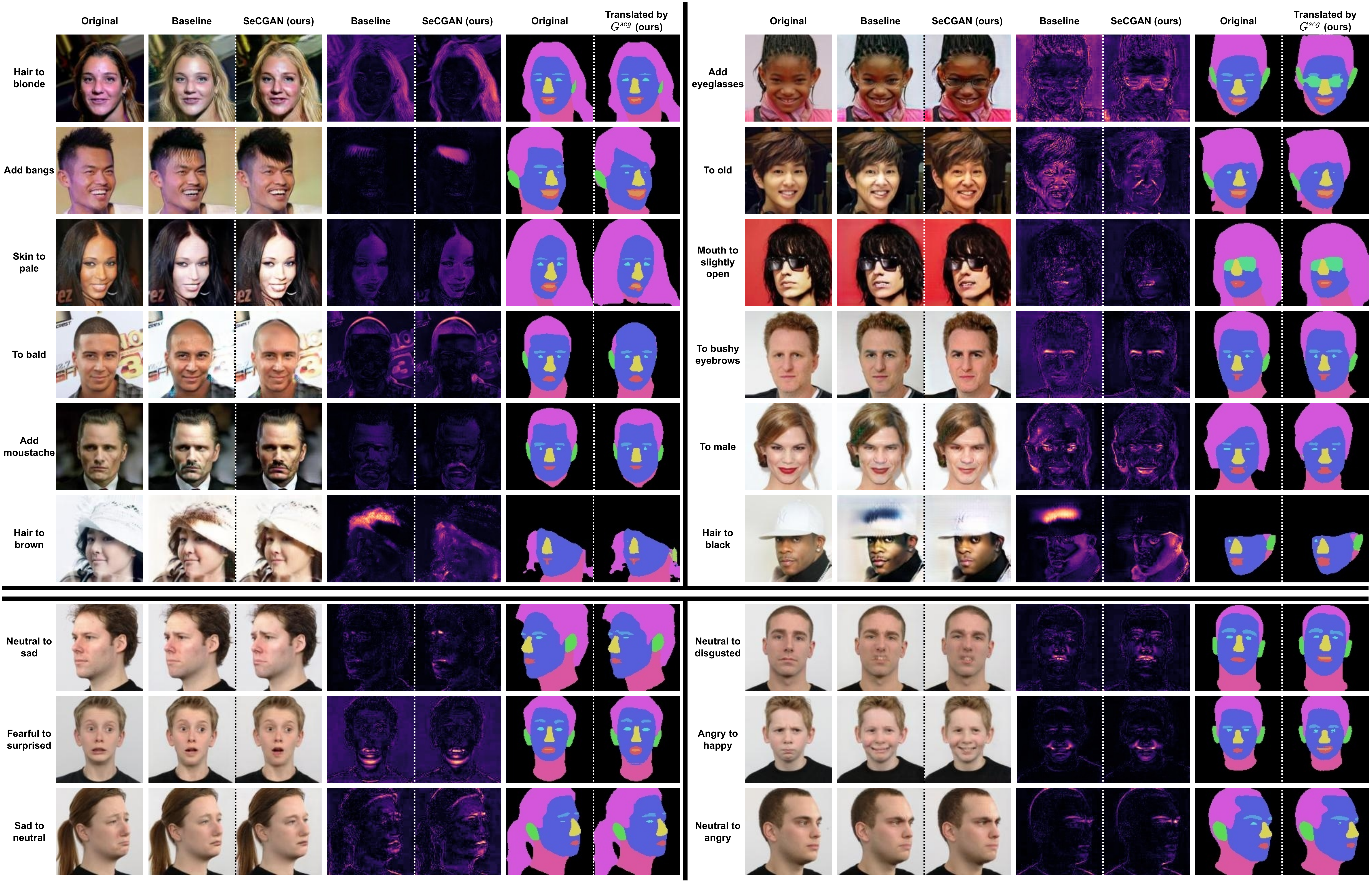}
    \caption{\textbf{Qualitative comparison between the baseline and our method on CelebA (top) and RaFD (bottom)} (zoom in for a better view). The first column are the original images. The second and third columns are images translated by the baseline and our method respectively. The fourth and fifth columns are heatmaps of the absolute difference between the original and translated images (the brighter, the larger the difference). The sixth and seventh columns are the original and translated semantic masks (by $G^{seg}$ in SeCGAN) respectively. More examples are included in the Appendix.}
    \label{fig:qualitative}
\end{figure*}

Figure \ref{fig:qualitative} gives visual comparisons between images generated by StarGAN (baseline) and SeCGAN (StarGAN backbone). On CelebA, it can be observed that our method produces more distinct attributes, particularly ones that correlate strongly with shapes of facial components such as ``add bangs'' and ``to bald''. As for other attributes which are only weakly correlated with shapes, the improvement is subtle but in fact observable via heatmaps of the absolute difference between translated and original images. More specifically, our method is better at focusing on regions that are relevant to the attribute in question as shown by the higher intensity of these regions in the heatmaps. For instance, this is reflected by the hair region in ``hair to blonde'', skin in ``skin to pale'', the area above the lips in ``add moustache'', the nasolabial folds and eyebags in ``to old'', and the eyebrows in ``to bushy eyebrows''. Similar observations can also be made from RaFD results. For example, our method makes more noticeable changes to the eyebrows in ``neutral to sad'', ``sad to neutral'', and ``neutral to angry''. Our method also adds more convincing teeth in ``neutral to disgusted'' and ``angry to happy'' whereas the teeth in the baseline appear to be only faintly pasted over the original image.

On the other hand, our method also fares better in avoiding unnecessary editing of unrelated regions. For instance, in ``hair to brown'' and ``hair to black'', the baseline is unable to locate the hair region in the input image correctly and instead directly pastes new colours over the hat, whereas our method is able to learn to ignore the hat region and focus on the hair. This can also be observed by comparing the background areas in examples including ``hair to blonde'' and ``to bald''. In addition, we also visualised the input semantic masks and the ones translated by the semantic generator $G^{seg}$. It can be observed that our method is able to correctly translate the semantic masks into the desired shapes for shape-related attributes/expressions, such as ``add bangs'', ``to bald'', ``add eyeglasses'', ``mouth to slightly open'', and ``neutral to sad'', whilst keeping the masks relatively intact for edits that are only weakly correlated with shapes, such as hair colour changes and ``add moustache''.
\section{Conclusion}
We present SeCGAN, a novel cGAN for face editing harnessing the benefits of higher-level semantic information and mutual learning. In training, SeCGAN uses two GANs for translating RGB images and semantic masks respectively whilst maintaining the semantic consistency of outputs. Experiments show that our method outperforms baselines across datasets in terms of accuracy whilst maintaining ssFID and IS. This improvement is also consistent in higher resolution settings or with fewer training data.

\textbf{Limitations and future work.} One limitation of our method is that it relies on a pre-trained parsing network, but our experiments show that pre-training could be done at a small scale and appears to work across datasets within the same domain. Our method also has twice the baseline’s
complexity but \emph{only} during training. Potential future directions include improving ssFID and IS alongside accuracy, and expanding current framework to other modalities, such as language, and domains, such as human body and animals. 

\textbf{Societal impact.} We acknowledge that our model is generative and thus can be used to create deep-fakes for disinformation, and we caution that appropriate detection algorithms should be implemented to guard against malicious use of deep-fakes. In fact, our model could be utilised as a data generator to train such detection algorithms.

\section{Acknowledgement}
This work was supported in part by the Croucher Foundation, Huawei Consumer Business Group, EPSRC Programme Grant ‘FACER2VM’ (EP/N007743/1), the Ministry of Land, Infrastructure and Transport of Korea / Korea Agency for Infrastructure Technology Advancement (22CTAP-C163793-02), and by National Research Council of Science and Technology funded by the Ministry of Science and ICT, Korea (CRC 21011).

{\small
\bibliographystyle{ieee_fullname}
\bibliography{egbib}
}

\appendix
\clearpage

\section{Appendix}
In this section, we present the algorithmic and architectural details of our method and also include additional qualitative examples.

\subsection{Algorithm}
Our detailed training procedure is shown in Algorithm \ref{alg:training}. The RGB and semantic branches are trained in an alternating manner. In other words, the semantic branch is frozen whilst the RGB branch is being updated and vice versa. 
\label{sec:algorithm}
\begin{algorithm}
\caption{SeCGAN.}
\label{alg:training}
	\begin{algorithmic}[1]
	    \STATE \textbf{Input:} Labelled dataset $\{(\mathbf{x}_1,\mathbf{y}_1),\ldots\}$ of RGB images and their corresponding attribute labels.
	    \STATE \textbf{Initialise:} Initial weights $\theta_{G^{rgb}}$, $\theta_{G^{seg}}$, $\theta_{D^{rgb}}$, and $\theta_{D^{seg}}$, parsing network $P$ with pre-trained weights, learning rates $\eta_G^t$ and $\eta_D^t$, step count $t=0$, number of discriminator updates per generator update $N$.
	    \REPEAT
	    \STATE $t\leftarrow t+1$
	    \STATE Randomly sample a mini-batch, and compute empirical loss terms $\mathcal{L}^{D^{rgb}}$ (\ref{eqn:d_rgb}) and $\mathcal{L}^{D^{seg}}$ (\ref{eqn:d_seg}).
	    \STATE Compute the gradients and update:
            \begin{align}
                \theta_{D^{rgb}} &\leftarrow \theta_{D^{rgb}} - \eta_D^t\frac{\partial \mathcal{L}^{D^{rgb}}}{\partial \theta_{D^{rgb}}}.\\
                \theta_{D^{seg}} &\leftarrow \theta_{D^{seg}} - \eta_D^t\frac{\partial \mathcal{L}^{D^{seg}}}{\partial \theta_{D^{seg}}}.
            \end{align}
        \IF{$t\equiv 0\mod{N}$}
        \STATE With the same mini-batch from line 5, compute empirical loss terms $\mathcal{L}^{G^{rgb}}$ (\ref{eqn:g_rgb}) and $\mathcal{L}^{G^{seg}}$ (\ref{eqn:g_seg}).
        \STATE Compute the gradients and update:
            \begin{align}
                \theta_{G^{rgb}} &\leftarrow \theta_{G^{rgb}} - \eta_G^t\frac{\partial \mathcal{L}^{G^{rgb}}}{\partial \theta_{G^{rgb}}}.\\
                \theta_{G^{seg}} &\leftarrow \theta_{G^{seg}} - \eta_G^t\frac{\partial \mathcal{L}^{G^{seg}}}{\partial \theta_{G^{seg}}}.
            \end{align}
        \ENDIF
	    \UNTIL{Convergence.}
	    \STATE \textbf{Output:} Optimal $G^{rgb}$.
	\end{algorithmic}
\end{algorithm}

\subsection{Architecture details}
In this subsection, we list all the layers of the generator $G^{seg}$ and discriminator $D^{seg}$ in the semantic branch. Here we present our approach implemented using both StarGAN (see Table \ref{tab:stargan_layers}) and AttGAN (see Table \ref{tab:attgan_layers}) as the backbone respectively. Here are some of the notations used: Conv($n$, $k$, $s$) means a convolutional layer with $n$ output channels, kernel size $k\times k$, and stride size $s$ (similar for ConvT which is a transposed convolutional layer); FC($c$) is a fully-connected layer with output dimension $c$; Res refers to a residual block; IN and BN refer to instance and batch normalisation respectively; LReLU is leaky ReLU; $n_a$ is the dimension of the attribute vector; $h$ is the height and width of the input images. The only differences separating $G^{seg}$ and $D^{seg}$ from their RGB counterparts $G^{rgb}$ and $D^{rgb}$ are the number of channels of the inputs (of both generator and discriminator), number of channels of the generator output, and the final activation function of the generator.

\begin{table}[t]
    \renewcommand{\arraystretch}{1.2}
    \centering
    \begin{tabularx}{\linewidth}{c|Y}
        \thickhline
        \multicolumn{2}{c}{\textbf{Generator $G^{seg}$}} \\ \thickhline
        \textit{Input shape} & \textit{Operations} \\ \hline
        $(\textbf{12}+n_a, h, h)$ & Conv(64, 7, 1), IN, ReLU \\ \hline
        $(64, h, h)$ & Conv(128, 4, 2), IN, ReLU \\ \hline
        $(128, \frac{h}{2}, \frac{h}{2})$ & Conv(256, 4, 2), IN, ReLU \\ \hline
        $(256, \frac{h}{4}, \frac{h}{4})$ & $6\times$ \{Res, Con(256, 3, 1), IN, ReLU\} \\ \hline
        $(256, \frac{h}{4}, \frac{h}{4})$ & ConvT(128, 4, 2), IN, ReLU \\ \hline
        $(128, \frac{h}{2}, \frac{h}{2})$ & ConvT(64, 4, 2), IN, ReLU \\ \hline
        $(64, h, h)$ & Conv(\textbf{12}, 7, 1, 3), Softmax \\
        \thickhline
        \multicolumn{2}{c}{\textbf{Discriminator $D^{seg}$}} \\ \thickhline
        \textit{Input shape} & \textit{Operations} \\ \hline
        $(\textbf{12}, h, h)$ & Conv(64, 4, 2), LReLU \\ \hline
        $(64, \frac{h}{2}, \frac{h}{2})$ & Conv(128, 4, 2), LReLU \\ \hline
        $(128, \frac{h}{4}, \frac{h}{4})$ & Conv(256, 4, 2), LReLU \\ \hline
        $(256, \frac{h}{8}, \frac{h}{8})$ & Conv(512, 4, 2), LReLU \\ \hline
        $(512, \frac{h}{16}, \frac{h}{16})$ & Conv(1024, 4, 2), LReLU \\ \hline
        $(1024, \frac{h}{32}, \frac{h}{32})$ & Conv(2048, 4, 2), LReLU \\ \hline
        \multirow{2}{*}{$(2048, \frac{h}{64}, \frac{h}{64})$} & $D_{adv}^{seg}$: Conv(1, 3, 1) \\ \hhline{~|-}
        & $D_{cls}^{seg}$: Conv($n_a$, $\frac{h}{64}$, 1), Sigmoid \\ \thickhline
    \end{tabularx}
    \caption{\textbf{Semantic branch architecture (backbone StarGAN).} The attribute label is concatenated with the input in the first layer, thus adding $n_a$ to the number of input channels.}
    \label{tab:stargan_layers}
\end{table}

\begin{table}[t]
    \renewcommand{\arraystretch}{1.2}
    \centering
    \begin{tabularx}{\linewidth}{c|Y}
        \thickhline
        \multicolumn{2}{c}{\textbf{Generator $G^{seg}$}} \\ \thickhline
        \textit{Input shape} & \textit{Operations} \\ \hline
        $(\textbf{12}, h, h)$ & Conv(64, 4, 2), BN, LReLU \\ \hline
        $(64, \frac{h}{2}, \frac{h}{2})$ & Conv(128, 4, 2), BN, LReLU \\ \hline
        $(128, \frac{h}{4}, \frac{h}{4})$ & Conv(256, 4, 2), BN, LReLU \\ \hline
        $(256, \frac{h}{8}, \frac{h}{8})$ & Conv(512, 4, 2), BN, LReLU \\ \hline
        $(512, \frac{h}{16}, \frac{h}{16})$** & Conv(1024, 4, 2), BN, LReLU \\ \hline
        $(1024+n_a, \frac{h}{32}, \frac{h}{32})$ & ConvT(1024, 4, 2), BN, ReLU \\ \hline
        $(1536+n_a, \frac{h}{16}, \frac{h}{16})$* & ConvT(512, 4, 2), BN, ReLU \\ \hline
        $(512, \frac{h}{8}, \frac{h}{8})$ & ConvT(256, 4, 2), BN, ReLU \\ \hline
        $(256, \frac{h}{4}, \frac{h}{4})$ & ConvT(128, 4, 2), BN, ReLU \\ \hline
        $(128, \frac{h}{2}, \frac{h}{2})$ & ConvT(\textbf{12}, 4, 2), Softmax \\
        \thickhline
        \multicolumn{2}{c}{\textbf{Discriminator $D^{seg}$}} \\ \thickhline
        \textit{Input shape} & \textit{Operations} \\ \hline
        $(\textbf{12}, h, h)$ & Conv(64, 4, 2), LReLU \\ \hline
        $(64, \frac{h}{2}, \frac{h}{2})$ & Conv(128, 4, 2), LReLU \\ \hline
        $(128, \frac{h}{4}, \frac{h}{4})$ & Conv(256, 4, 2), LReLU \\ \hline
        $(256, \frac{h}{8}, \frac{h}{8})$ & Conv(512, 4, 2), LReLU \\ \hline
        $(512, \frac{h}{16}, \frac{h}{16})$ & Conv(1024, 4, 2), LReLU \\ \hline
        \multirow{2}{*}{$(1024, \frac{h}{32}, \frac{h}{32})$} & $D_{adv}^{seg}$: FC(1024), ReLU, FC(1) \\ \hhline{~|-}
        & $D_{cls}^{seg}$: FC(1024), ReLU, FC($n_a$), Sigmoid \\ \thickhline
    \end{tabularx}
    \caption{\textbf{Semantic branch architecture (backbone AttGAN).} Unlike StarGAN, the attribute label is injected into the generator decoder rather than at the very beginning. The additional channels in the input * also come from a skip connection which concatenates the current input with the previous output **. Combining both leads to the $512+n_a$ channels in addition to the 1024 from the previous output layer.}
    \label{tab:attgan_layers}
\end{table}

\subsection{Additional qualitative results}
\begin{figure*}
    \centering
    \includegraphics[width=\textwidth]{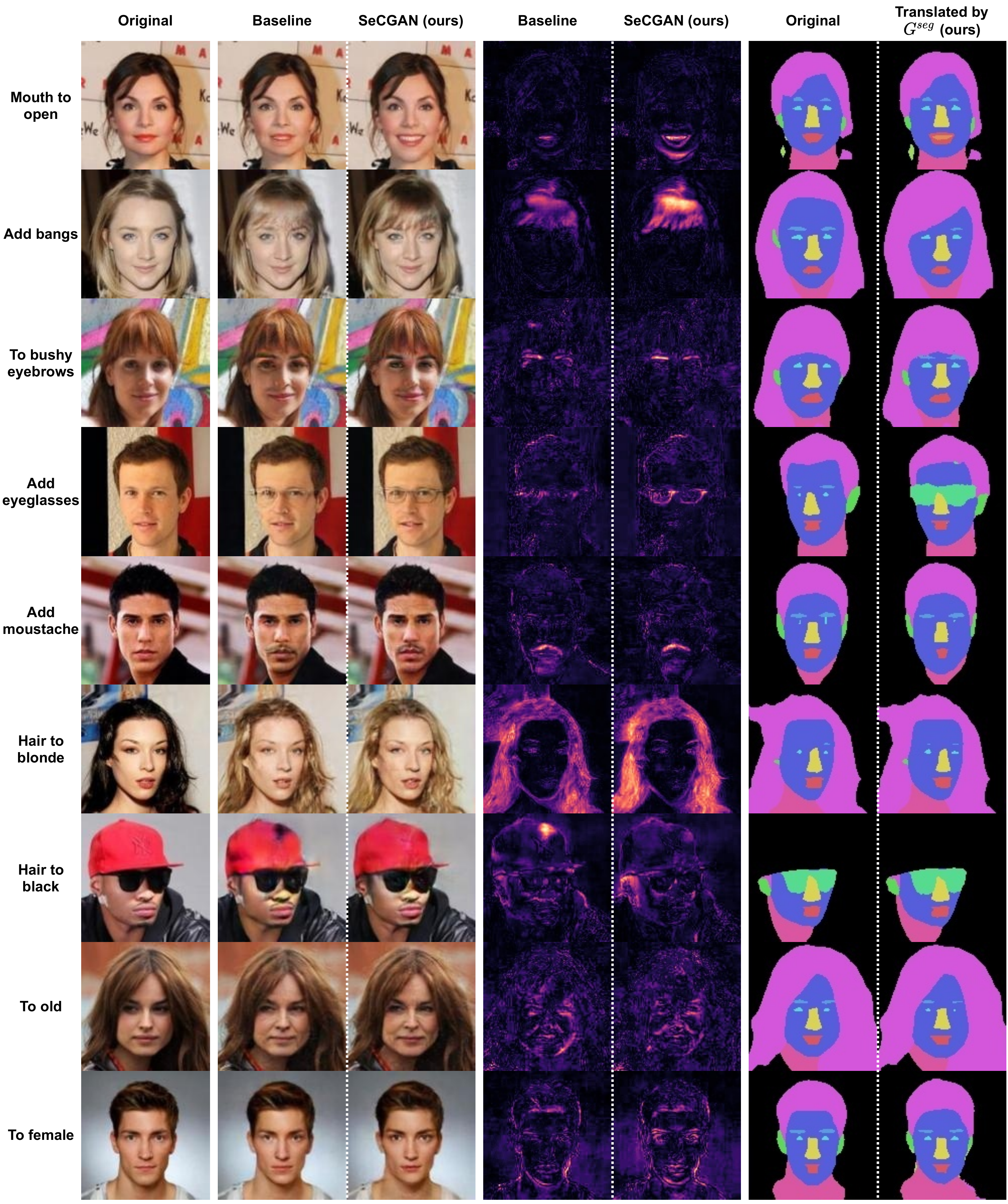}
    \caption{\textbf{Qualitative comparison between the baseline and our method (AttGAN backbone) at $\mathbf{128\times 128}$}. The columns are ordered in the same way as those in Figure \ref{fig:qualitative}. Similar to Figure \ref{fig:qualitative}, our method can be observed to produce more focused and precise translations.}
    \label{fig:qualitative_attgan}
\end{figure*}

\begin{figure*}
    \centering
    \includegraphics[width=\textwidth]{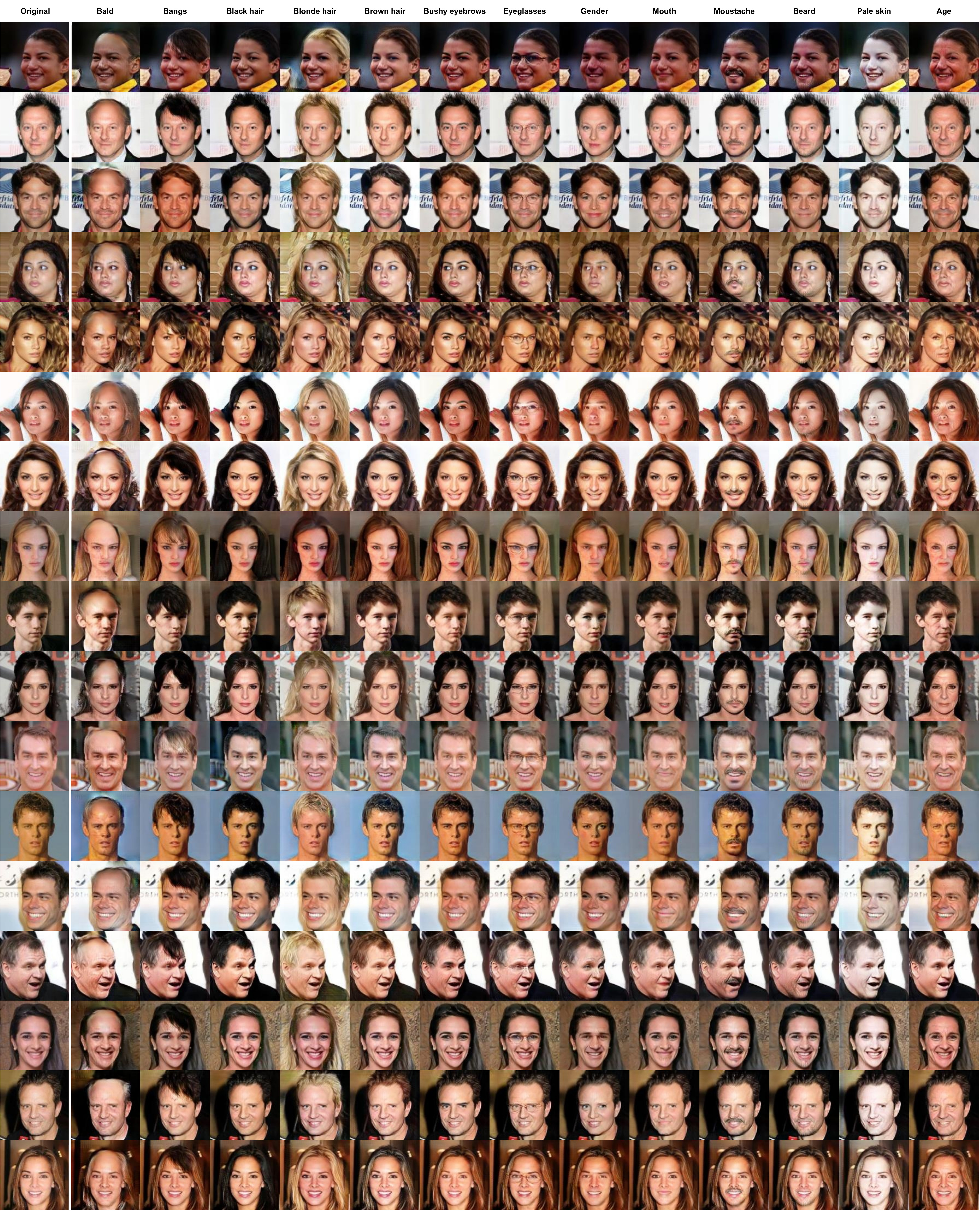}
    \caption{\textbf{Qualitative results of our method (StarGAN backbone) at $\mathbf{128\times 128}$} (zoom in for a better view).}
    \label{fig:qualitative_2}
\end{figure*}

\begin{figure*}
    \centering
    \includegraphics[width=\textwidth]{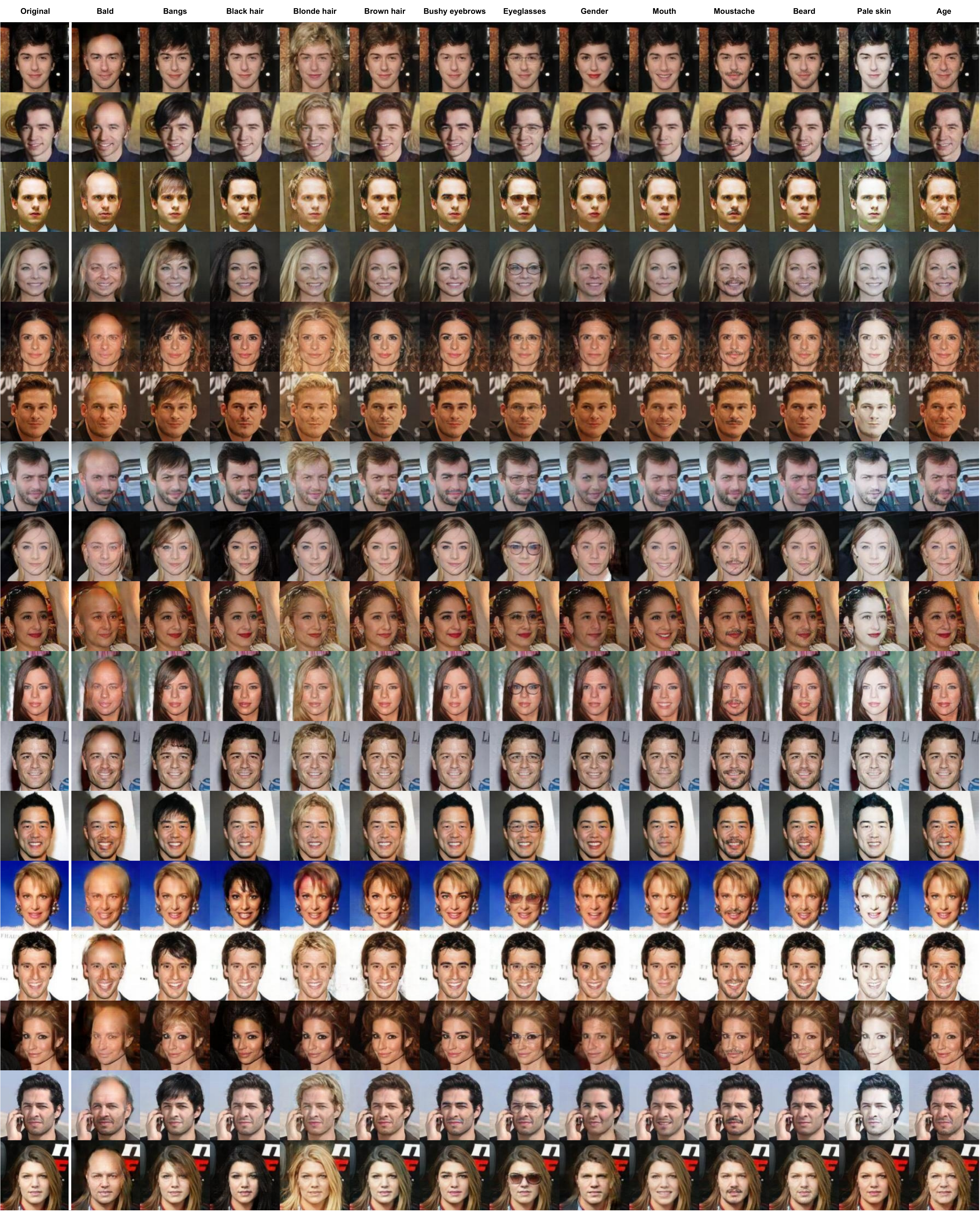}
    \caption{\textbf{Qualitative results of our method (AttGAN backbone) at $\mathbf{128\times 128}$} (zoom in for a better view).}
    \label{fig:qualitative_3}
\end{figure*}

Similarly to Figure \ref{fig:qualitative}, we also provide a qualitative comparison using AttGAN as the backbone in Figure \ref{fig:qualitative_attgan}. Again, we can observe that our method produces more focused and accurate translations than baseline, such as in ``mouth to open'', ``add bangs'', and ``to bushy eyebrows''. We also include additional qualitative results of SeCGAN trained on CelebA at resolution $128\times 128$, implemented using StarGAN (see Figure \ref{fig:qualitative_2}) and AttGAN (see Figure \ref{fig:qualitative_3}) as backbones respectively. In these figures, each translated image represents an attribute reversal. In the ``gender'' column, for example, the gender of the original image is reversed. This also applies to the three hair colours, but negating one hair colour does not mean the target hair colour must be one of the other two. For instance, reversing ``blonde hair'' on a person who already has blonde hair would simply result in a neutral hair colour rather than brown or black.

\clearpage

\end{document}